\documentclass[letterpaper, 10 pt, conference]{ieeeconf}
\IEEEoverridecommandlockouts                                       
\usepackage{cite}
\usepackage{algorithm, algpseudocode}  %algorithmic}
\usepackage{graphics} 
\usepackage{epsfig} % for postscript graphics files
\usepackage{amssymb} % assumes amsmath package installed
\usepackage{amsmath} % assumes amsmath package installed
\usepackage{latexsym}
\usepackage{mathrsfs} %
\usepackage{xcolor}
\usepackage{cases}
\usepackage[normalem]{ulem}
\usepackage{bm}       % Helps with making symbols bold in math mode
\usepackage{graphicx}
\usepackage{verbatim}
\usepackage{balance}
\usepackage{geometry} 
\usepackage{booktabs}% http://ctan.org/pkg/booktabs
\usepackage[colorlinks=false,hidelinks]{hyperref} % For hyperlinks in the PDF
\usepackage{url}
\usepackage{environ}
\usepackage{noel}
\usepackage{caption}
\usepackage{subcaption}
\usepackage{mathtools}

\NewEnviron{ruledtable}{%
\par\addvspace{2mm}\hrule height 0.03cm 
\begin{table}[!h]\BODY\end{table}
\hrule height 0.03cm \addvspace{2mm}
}

\title{\LARGE \bf
Learning Controller Gains on Bipedal Walking Robots \\ via User Preferences

}

\author{Noel Csomay-Shanklin$^{1}$, Maegan Tucker$^{2}$, Min Dai$^{2}$, Jenna Reher$^{2}$, Aaron D. Ames$^{1,2}$% <-this % stops a space
\thanks{This research was supported by NSF NRI award 1924526, NSF award 1932091, NSF CMMI award 1923239, NSF Graduate Research Fellowship No. DGE‐1745301, and the Caltech Big Ideas and ZEITLIN Funds.}%
\thanks{$^{1}$Authors are with the Department
of Computing and Mathematical Sciences, California Institute of Technology,
Pasadena, CA 91125.}%
\thanks{$^{2}$Authors are with the Department
of Mechanical and Civil Engineering, California Institute of Technology,
Pasadena, CA 91125.}
}

\begin{document}

\maketitle
\thispagestyle{empty}
\pagestyle{empty}

%%%%%%%%%%%%%%%%%%%%%%%%%%%%%%%%%%%%%%%%%%%%%%%%%%%%%%%%%%%%%%%%%%%%%%%%%%%%%%%%
\begin{abstract}
% Experimental demonstration of complex robotic behaviors relies heavily on finding the correct controller gains. This painstaking process is often completed by a domain expert, requiring deep knowledge of the relationship between parameter values and the resulting behavior of the system. Even when such knowledge is possessed, it can take significant effort to navigate the unintuitive landscape of possible parameter combinations. In this work, we explore the extent to which preference-based learning can be used to optimize controller gains online by repeatedly querying the user for their preferences. This methodology is applied to two variants of the control Lyapunov function based nonlinear controllers framed as quadratic programs --- a class of nonlinear optimization-based controllers that have nice theoretic properties, yet are challenging to realize in practice due to nonintuitive parameter tuning. In this work, we experimentally demonstrate these control variants both on the planar underactuated biped, AMBER, and on the 3D underactuated biped, Cassie. We give background on the control and learning frameworks, show that the proposed method is repeatably able to learn comparable gains to a domain expert in a matter of hours, and experimentally demonstrate the performance and robustness of the learned controllers. 

Experimental demonstration of complex robotic behaviors relies heavily on finding the correct controller gains. This painstaking process is often completed by a domain expert, requiring deep knowledge of the relationship between parameter values and the resulting behavior of the system. Even when such knowledge is possessed, it can take significant effort to navigate the nonintuitive landscape of possible parameter combinations. In this work, we explore the extent to which preference-based learning can be used to optimize controller gains online by repeatedly querying the user for their preferences. This general methodology is applied to two variants of control Lyapunov function based nonlinear controllers framed as quadratic programs, which provide theoretical guarantees but are challenging to realize in practice. These controllers are successfully demonstrated both on the planar underactuated biped, AMBER, and on the 3D underactuated biped, Cassie. We experimentally evaluate the performance of the learned controllers and show that the proposed method is repeatably able to learn gains that yield stable and robust locomotion.

\end{abstract}

% ****************** SECTIONS *******************

%%%%%%%%%%%%%%%%%%%%%%%%%%%%%%%%%%%%%%%%%%%%%%%%%%%%%%%%%%%%%%%%%%%%%%%%%%%%%%%%
\section{Introduction} \label{sec:intro}

Achieving robust and stable performance for physical robotic systems relies heavily on careful gain tuning, regardless of the implemented controller. Navigating the space of possible parameter combinations is a challenging endeavor, even for domain experts. To combat this challenge, researchers have developed systematic ways to tune gains for specific controller types \cite{PIDtuning,MPCtuning,zheng1992practical,hjalmarsson1998iterative}.
% \cite{PIDtuning,MPCtuning,zheng1992practical, zhao2012performance, hjalmarsson1998iterative, sung1996limitations, odgaard2016using}
For controllers where the input/output relationship between parameters and the resulting behavior is less clear, this can be prohibitively difficult. These difficulties are especially prevalent in the setting of bipedal locomotion, due to the extreme sensitivity of the stability of the system with respect to controller gains.

\begin{figure}[!t]
    \centering
    \includegraphics[width=0.923\linewidth]{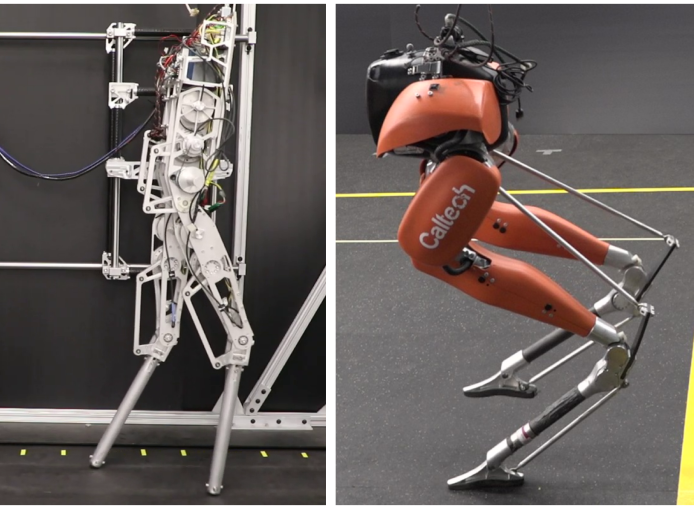} 
    % \hfill
    % \includegraphics[width=0.45\linewidth]{Figures/CassieAnimated2.png}
    \caption{The two experimental platforms investigated in this work: the planar AMBER-3M point-foot robot \cite{ambrose2017toward} (left), and the 3D Cassie robot \cite{cassie} (right).}
    \label{fig:cassie}
\end{figure}

% Stable walking can be achieved through the construction of stabilizing controllers.
It was shown in \cite{ames2013towards} that control Lyapunov functions (CLFs) are capable of stabilizing locomotion through the hybrid zero dynamics (HZD) framework, with \cite{galloway2015torque} demonstrating how this can be implemented as a quadratic program (QP), allowing the problem to be solved in a pointwise-optimal fashion even in the face of feasibility constraints. However, achieving robust walking behavior on physical bipeds can be an arduous process due to complexities such as compliance, under-actuation, and narrow domains of attraction. One such controller that has recently demonstrated stable locomotion on the 22 degree of freedom (DOF) Cassie biped, as shown in Fig. \ref{fig:cassie}, is the ID-CLF-QP$^+$ \cite{jenna2021tro}.
% , a variant of the CLF-QP that will be utilized in this work.

%

% \begin{figure*}[t!]
% \centering
%      \begin{subfigure}[b]{0.47\textwidth}
% \centering
% 	\vspace{2mm}
% 	\includegraphics[width= 1.2\columnwidth]{Figures/floating_base_coordinate.pdf}
% % 	\caption{\red{The configuration coordinates of the Cassie robot: with the side and front views highlighting the compliant mechanism and the general morphology of the robot.}
%     % }
% 	\label{fig:cassie_configuration}
%      \end{subfigure}
%      \hfill
%      \begin{subfigure}[b]{0.15\textwidth}
% \centering
% 	\vspace{2mm}
% 	\includegraphics[width= 1\columnwidth]{Figures/AmberConfig.pdf}
% % 	\caption{\red{The configuration coordinates of the Cassie robot: with the side and front views highlighting the compliant mechanism and the general morphology of the robot.}
% %     }
% 	\label{fig:cassie_configuration}
%      \end{subfigure}
%          \caption{\textit{Left}: Configuration of the 22 DOF Cassie robot\cite{cassie}. \textit{Right}: Config. of the 7 DOF planar robot AMBER-3M\cite{ambrose2017toward}.}
%          \label{fig:Configs}
%          \vspace{-5mm}
% \end{figure*}

%
% However, 
Synthesizing a controller capable of accounting for the challenges of underactuated locomotion, such as the ID-CLF-QP$^+$, necessitates the addition of numerous control parameters, exacerbating the issue of gain tuning. Moreover, the relationship between the control parameters and the resulting behavior of the robot is extremely nonintuitive and results in a landscape that requires dedicated time to navigate, even for domain experts. 
% For example, the implementation of the ID-CLF-QP$^+$ in \cite{jenna2021tro} entailed 2 dedicated months of hand-tuning around 60 control parameters.
Recently, machine learning techniques have been implemented to alleviate the process of hand-tuning gains in a controller agnostic way by systematically navigating the entire parameter space \cite{birattari2009tuning,jun1999automatic,marco2016automatic}. More specifically, Bayesian optimization techniques have been applied to learning gait parameters and controller gains for various bipedal systems \cite{calandra2016bayesian,ATRIASpbl}. However, these techniques rely on a carefully constructed predefined reward function.
Furthermore, it is often the case that different desired properties of the robotic behavior are conflicting and therefore can't be simultaneously optimized. 

To alleviate the gain tuning process and enable the use of complicated controllers for na\"ive users, we employ a preference-based learning framework that only relies on subjective user feedback, mainly pairwise preferences, to systematically search the parameter space and realize stable and robust experimental walking. Preferences are a particularly useful feedback mechanism for parameter tuning because they are able to capture the notion of ``general goodness'' without a predefined reward function. This is particularly important for bipedal locomotion due to the lack of commonly agreed upon numerical metric of good or even stable walking in the community \cite{wieber2002stability,vukobratovic2004zero,pratt2006velocity,grizzle2014models}. 

Preference-based learning has been previously used towards selecting essential constraints of an HZD gait generation framework which resulted in stable and robust experimental walking on a planar biped with unmodeled compliance at the ankle \cite{tucker2020preferencebased}. In this paper, we build on the previous work by exploring the application of preference-based learning towards implementing optimization-based controllers on multiple bipedal platforms. Specifically, we demonstrate the framework towards tuning gains of a CLF-QP$^{+}$ controller on the AMBER bipedal robot, as well as an ID-CLF-QP$^{+}$ controller on the Cassie bipedal robot, requiring the learning framework to operate in a much higher-dimensional space.

\section{Preliminaries on Dynamics and Control} \label{sec:opt}
\subsection{Modeling and Gait Generation}

% \begin{figure}[t]
% 	\centering
% 	\vspace{2mm}
% 	\includegraphics[width= 1\columnwidth]{Figures/floating_base_coordinate.pdf}
% 	\caption{\red{The configuration coordinates of the Cassie robot: with the side and front views highlighting the compliant mechanism and the general morphology of the robot.}
%     }
% 	\label{fig:cassie_configuration}
% \end{figure}
%
\begin{figure}
    \centering
    \includegraphics[width=\linewidth]{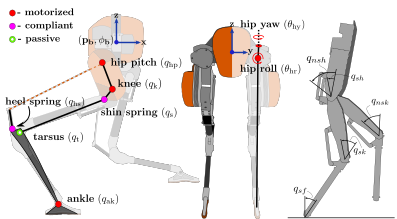}
    \caption{Configuration of the 22 DOF (using an unpinned model) Cassie robot\cite{cassie} (left) and configuration of the 5 DOF (using a pinned model) planar robot AMBER-3M\cite{ambrose2017toward} (right).}
    \label{fig:Configs}
\end{figure}

Following a floating-base convention \cite{grizzle2014models}, we define the configuration space as $\mathcal{Q} \subset\R^n$, where $n$ is the unconstrained DOF (degrees of freedom). Let $q = (p_b, \phi_b, q_l) \in \mathcal{Q}:= \R^3\times SO(3)\times \mathcal{Q}_l,$
where $p_b$ is the global Cartesian position of the body fixed frame attached to the base linkage (the pelvis), $\phi_b$ is its global orientation, and $q_l\in\mathcal{Q}_l\subset\R^{n_l}$ are the local coordinates representing rotational joint angles. Further, the state space $\mathcal{X}=T\mathcal{Q}\subset \R^{2n}$ has coordinates $x=(q^\top,\dot q^\top)^\top$. 
%Take $q\in \mathcal{Q}\subset\mathbb{R}^n$ to be the configuration space of the robot, with the full state given by $x=(q,\dot{q})\in T\mathcal{Q}$. 
The robot is subject to various \textit{holonomic constraints}, which can be summarized by an equality constraint $h(q)\equiv0$ where $h(q)\in\R^h$.  
Differentiating $h(q)$ twice and applying D'Alembert's principle to the Euler-Lagrange equations for the constrained system, the dynamics can be written as:
\begin{align}
    D(q)\ddot{q} + H(q,\dot{q}) &= Bu + J(q)^\top \lambda \label{dyn} \\
    J(q)\ddot{q} + \dot{J}(q,\dot{q})\dot{q} &= 0 \label{holconstr}
\end{align}
where $D(q)\in\R^{n\times n}$ is the mass-inertia matrix, $H(q,\dot{q})$ contains the Coriolis, gravity, and additional non-conservative forces, $B\in\R^{n\times m}$ is the actuation matrix, $J(q)\in\R^{h\times n}$ is the Jacobian matrix of the holonomic constraint, and $\lambda\in\R^h$ is the constraint wrench. The system of equations \eqref{dyn} for the dynamics can also be written in control-affine form:
\begin{align*}
    \dot{x} = \underbrace{\begin{bmatrix}\dot{q}\\-D(q)^{-1}(H(q,\dot q)-J(q)^\top \lambda)\end{bmatrix}}_{f(x)} + \underbrace{\begin{bmatrix}0\\D(q)^{-1}B\end{bmatrix}}_{g(x)}u.
\end{align*}
The mappings $f:T\mathcal{Q}\to \R^n$ and $g:T\mathcal{Q}\to \R^{n\times m}$ are assumed to be locally Lipschitz continuous. 

Dynamic and underactuated walking consists of periods of continuous motion followed by discrete impacts, which can be accurately modeled within a hybrid framework \cite{westervelt2018feedback}. If we consider a bipedal robot undergoing domains of motion with only one foot in contact (either the left ($L$) or right ($R$)), and domain transition triggered at footstrike, then we can define:
\begin{align*}
    \mathcal{D}_{SS}^{\{L,R\}} &= \{(q,\dot q):p_{swf}^z(q) \ge 0\}, \\
    \mathcal{S}_{L\to R, R\to L} &= \{(q,\dot q):p_{swf}^z(q) = 0, \dot p_{swf}^z(q,\dot q) < 0\},
\end{align*}
where $p_{swf}^z:\mathcal{Q}\to\R$ is the vertical position of the swing foot, and $\mathcal{D}_{SS}^{\{L,R\}}$ is the continuous domain on which our dynamics \eqref{dyn} evolve with a transition from one stance leg to the next triggered by the switching surface $\mathcal{S}_{L\to R, R\to L}$.
When this domain transition is triggered, the robot undergoes an impact with the ground, yielding a hybrid model:
\begin{align}
    \mathcal{HC} = \begin{cases} 
         \dot{x} = f(x) + g(x) u &x\not\in\mathcal{S}_{L\to R, R\to L} \\
         \dot{x}^+ = \Delta(x^-) &x\in\mathcal{S}_{L\to R, R\to L}
    \end{cases}\label{hybridcontrolsystem}
\end{align}
where $\Delta$ is a plastic impact model \cite{grizzle2014models} applied to the pre-impact states, $x^-$, such that the post-impact states, $x^+$, respect the holonomic constraints of the subsequent domain. 

In this work, we design locomotion using the \textit{hybrid zero dynamics} (HZD) framework \cite{westervelt2018feedback} in order to generate stable periodic walking for underactuated bipeds. At the core of this method is the regulation of \textit{virtual constraints}, or outputs:
\begin{align}
    y(x,\tau,\alpha) &= y_a(x)-y_d(\tau,\alpha), \label{outputeqs}
\end{align}
with the goal of driving $y\to0$ where $y_a:T\mathcal{Q}\to\R^p$ and $y_d:T\mathcal{Q}\times \R\times\R^a\to\R^p$ are smooth functions representing the actual and desired outputs, respectively, $\tau$ is a phasing variable, and $\alpha$ is a set of Bezi\`er polynomial coefficients that can be shaped to encode stable locomotion.

The desired outputs were optimized using the FROST toolbox \cite{FROST}, where stability of the gait was ensured in the sense of Poincar\'e via HZD theory \cite{westervelt2003hybrid}. This was done first for AMBER, in which one walking gait was designed using a pinned model of the robot\cite{ambrose2017toward}, and then on Cassie for $3$D locomotion using the motion library found in \cite{GaitLibrary} consisting of $171$ walking gaits for speeds in $0.1$ m/s intervals on a grid for sagittal speeds of $v_x\in[-0.6,1.2]$ m/s and coronal speeds of $v_y\in[-0.4,0.4]$ m/s.

% In this setting, the cost function is taken to be 
% \begin{align}
%     \mathcal{J}(\mathbf{X}) = \int_0^{t_f}c_u|u|^2+c_\phi |\phi_b|^2 dt
% \end{align}
% where $\phi_b$ is the global orientation of the body frame, and $c_{(\cdot)}$ are weights. Here, $c_u=0.0001$ and $c_\phi=(20,1,30)$.
%
%
%
\subsection{Control Lyapunov Functions}
Control Lyapunov functions (CLFs), and specifically rapidly exponentially stabilizing control Lyapunov functions (RES-CLFs), were introduced as methods for achieving (rapidly) exponential stability on walking robots \cite{RES-CLF}. 
This control approach has the benefit of yielding a control framework that can provably stabilize periodic orbits for hybrid system models of walking robots, and can be realized in a pointwise optimal fashion. 
%In this work, Control Lyapunov functions (CLFs) are used for the purpose of trajectory tracking. 
In this work, we consider only outputs which are \textit{vector relative degree} $2$. Thus, differentiating \eqref{outputeqs} twice with respect to the dynamics results in:
\begin{align*}
     \ddot y(x) = L_f^2 y(x) + L_gL_fy(x)u.
\end{align*}
where $L_f^2y(x):T\mathcal{Q}\to\mathbb{R}^p$ and $L_gL_fy(x):T\mathcal{Q}\to\mathbb{R}^p$ represent the Lie derivatives of the outputs with respect to the vector fields $f(x)$ and $g(x)$.
Assuming that the system is feedback linearizeable, we can invert the decoupling matrix, $L_gL_fy(x)$, to construct a preliminary control input:
\begin{align}
    u = \left( L_gL_fy(x)\right)^{-1}\left(\nu-L_f^2 y(x) \right), \label{FL}
\end{align}
which renders the output dynamics to be $\ddot y=\nu$. With the auxiliary input $\nu$ appropriately chosen, the nonlinear system can be made exponentially stable. Assuming the preliminary controller \eqref{FL} has been applied to our system, and defining $\eta=[y,\dot y]^\top$ we have the following output dynamics \cite{Isidori95}:
\begin{align}
    \dot{\eta} = \underbrace{\begin{bmatrix}0&I\\0&0\end{bmatrix}}_{F}\eta + \underbrace{\begin{bmatrix}0\\I\end{bmatrix}}_Gv. \label{linearoutput}
\end{align}
%This methodology can be extended to feedback linearizable systems of arbitrary relative degree \cite{Isidori95}. 
With the goal of constructing a CLF using \eqref{linearoutput}, we evaluate the continuous time algebraic Ricatti equation (CARE):
\begin{align}
    F^\top P + PF + PGR^{-1}G^\top P + Q = 0, \tag{CARE} \label{CARE}
\end{align}
which has a solution $P \succ 0$ for any $Q=Q^\top\succ 0$ and $R=R^\top\succ 0$. 
%There is no closed form solution to \eqref{CARE}; instead, solving it requires iteratively updating an estimate of $P$ until convergence is attained. As such, there is little intuition for how changing elements of $Q$ will impact the resulting solution. 
%
From the solution of \eqref{CARE}, we can construct a rapidly exponentially stabilizing CLF (RES-CLF) \cite{RES-CLF}:
\begin{align}
    V(\eta) = \eta ^\top I_\epsilon P I_\epsilon \eta, \hspace{5mm} I_\epsilon = \begin{bmatrix}\frac{1}{\epsilon}I&0\\0&I\end{bmatrix},
\end{align}
where $0<\epsilon<1$ is a tunable parameter that drives the (rapidly) exponential convergence. 
Any feedback controller, $u$, which can satisfy the convergence condition:
\begin{align}
    \dot{V}(\eta) &= L_fV(\eta) + L_gV(\eta) u \le - \frac{1}{\epsilon}\underbrace{\frac{\lambda_{min}(Q)}{\lambda_{max}(P)}}_{\gamma} V(\eta),\label{eq::V_dot}
\end{align}
will then render rapidly exponential stability for the output dynamics \eqref{outputeqs}. To enforce \eqref{eq::V_dot}, a quadratic program (CLF-QP) \cite{galloway2015torque}, with \eqref{eq::V_dot} as an inequality constraint can be posed.

% In the context of RES-CLF, we can then define:
% \begin{align*}
%     K_\epsilon(x) = \{u_\epsilon \in U : L_f V(x) + L_g V(x) u + \frac{\gamma}{\epsilon} V(x) \leq 0 \},
% \end{align*}
% describing an entire class of the controllers which result in (rapidly) exponential convergence. 

% , a weighted relaxation term, $\delta$, is added \eqref{eq::V_dot} in order to maintain feasibility.

% \begin{ruledtable}
% % \vspace{-2mm}
% {\textbf{\normalsize CLF-QP-$\delta$:}}
% \par\vspace{-4mm}{\small 
% \begin{align}
%         \label{eq:CLF-QP-d}
% 	{u}^* = \underset{u\in\mathbb{R}^m}{\mathrm{argmin}} &\hspace{3mm} \|L_f^2y(x) + L_gL_fy(x)u \|^2 + w_{\dot V}  \delta^2 \\
% 	\mathrm{s.t.} 		&\hspace{3mm} \dot V(x) = L_fV(x) + L_gV(x)u \le -\frac{\gamma}{\epsilon} V + \delta \notag \\
% 	                    &\hspace{3mm}  u_{min}\preceq u \preceq u_{max}  \notag 
% \end{align}}\vspace{-4mm}\par
% \end{ruledtable}
% \noindent Because this relaxation term is penalized in the cost, we could also move the inequality constraint completely into the cost as an exact penalty function \cite{jenna2021tro}:
% \begin{align*}
%     \mathcal{J}_{\mathrm{\delta}} = \|L_f^2y(x) + L_gL_fy(x)u \|^2 + w_{\dot V} || g^+(x,u) ||
% \end{align*}
% where:
% \begin{align*}
%         g(x,u) &:= L_f V(x) + L_g V(x) u + \frac{\gamma}{\epsilon} V(x), \\
%         g^+(x,u) &\triangleq \max(g,0), 
% \end{align*}
% One of the downsides to using this approach is that the cost term $|| g^+(x,u) ||$ will intermittently trigger and cause a jump to occur in the commanded torque. 

Implementing this controller on physical systems, which are often subject to additional constraints such as torque bounds or friction limits, suggests that relaxation for the inequality constraint should be used. The introduction of relaxation and the need to reduce torque chatter on physical hardware lead to the following relaxed (CLF-QP) with incentivized convergence in the cost \cite{reher2020inverse}:

%
% \begin{ruledtable}
% % \vspace{-2mm}
% {\textbf{\normalsize CLF-QP:}}
% \par\vspace{-4mm}{\small 
% \begin{align}
%         \label{CLF-QP}
% 	{u}^* = \underset{u\in\mathbb{R}^m}{\mathrm{argmin}} &\hspace{3mm} \|L_f^2y(x) + L_gL_fy^{-1}(x)u \|^2 \\
% 	\mathrm{s.t.} 		&\hspace{3mm}  L_fV(x,\alpha) + L_gV(x,\alpha)u \le -\gamma V(x,\alpha)  \notag 
% \end{align}}\vspace{-4mm}\par
% 
% \end{ruledtable}
%
\begin{ruledtable}
% \vspace{-2mm}
{\textbf{\normalsize CLF-QP$^+$:}}
\par\vspace{-4mm}{\small 
\begin{align}
        \label{eq:CLF-QP-p}
	{u}^* = \underset{u\in\mathbb{R}^m}{\mathrm{argmin}} &\hspace{3mm} \|L_f^2y(x) + L_gL_fy(x)u \|^2 + w_{\dot{V}}\dot V(x, u) \\
	\mathrm{s.t.} 		&\hspace{3mm}  u_{min}\preceq u \preceq u_{max}  \notag 
\end{align}}\vspace{-4mm}\par

\end{ruledtable}

In order to avoid computationally expensive inversions of the model sensitive mass-inertia matrix, and to allow for a variety of costs and constraints to be implemented, a variant of the (CLF-QP) termed the (ID-CLF-QP) was introduced in \cite{reher2020inverse}. This controller is used on the Cassie biped, with the decision variables $\mathcal{X}=[\ddot q^\top, u^\top, \lambda^\top]^\top\in\R^{39}$:
\newpage
\begin{ruledtable}
\vspace{-2mm}
{\textbf{\normalsize ID-CLF-QP$^+$:}}
\par\vspace{-4mm}{\small 
\begin{align}
        \label{ID-CLF-QP}
	\mathcal{X}^* = \underset{\mathcal{X}\in\mathbb{X}_{ext}}{\mathrm{argmin}} &\hspace{3mm} \|A(x)\mathcal{X} - b(x)\|^2 +\dot{V}(q,\dot q,\ddot q) \\
	\mathrm{s.t.} 		&\hspace{3mm}  D(q)\ddot{q} + H(q,\dot{q}) = Bu + J(q)^\top \lambda  \notag \\
				  		&\hspace{3mm}  u_{min}\preceq u \preceq u_{max}  \notag \\
				  		&\hspace{3mm} \lambda \in \mathcal{AC}(\mathcal{X})
\end{align}}\vspace{-6mm}\par
\end{ruledtable}
\noindent where \eqref{holconstr} has been moved into the cost terms $A(x)$ and $b(x)$ as a weighted soft constraint, in addition to a feedback linearizing cost, and a regularization for the nominal $\mathcal{X}^*(\tau)$ from the HZD optimization.
Interested readers are referred to \cite{jenna2021tro, reher2020inverse} for the full (ID-CLF-QP+) formulation.

\begin{table}[b!]
\centering
\caption{Learned Parameters}
\label{table:bounds}
\def\arraystretch{1.2} % 1 is the default, increase for more row spacing
        \begin{tabular}{ | l | c | c|} 
        \hline
        \multicolumn{3}{|c|}{\textsc{CASSIE}} \\
        \hline
         & Pos. Bounds & Vel. Bounds \\
        \hline
        $Q$ Pelvis Roll ($\phi_x$)  &  $a_1$:[2000, 12000] & $a_7$:[5, 200] \\ 
        \hline
        $Q$ Pelvis Pitch ($\phi_y$) & $a_2$:[2000, 12000] & $a_8$:[5, 200]\\
        \hline
        $Q$ Stance Leg Length ($\|\phi^{st}\|_2$) &  $a_3$:[4000, 15000] & $a_9$:[50, 500] \\ 
        \hline
        $Q$ Swing Leg Length ($\|\phi^{sw}\|_2$)  & $a_4$:[4000, 20000] & $a_{10}$:[50, 500] \\ 
        \hline
        $Q$ Swing Leg Angle ($\theta_{hp}^{sw}$)  & $a_5$:[1000, 10000] & $a_{11}$:[10, 200] \\
        \hline
        $Q$ Swing Leg Roll ($\theta_{hr}^{sw}$)  &  $a_6$:[1000, 8000] & $a_{12}$:[5, 150] \\ 
        \hline
    \end{tabular}
\end{table}

\begin{table}[b!]
\vspace{-3mm}
\centering
\def\arraystretch{1.18} % 1 is the default, increase for more row spacing
        \begin{tabular}{ | l  | c | c|| l | c |} 
        \hline
        \multicolumn{5}{|c|}{\textsc{AMBER}} \\
        \hline
         & Pos. Bounds & Vel. Bounds & & Bounds\\
        \hline
        $Q$ Knees  &  $a_{1}$:[100, 1500] & $a_{3}$:[10, 300] & $\epsilon$ & $a_{5}$:[0.08, 0.2] \\ 
        \hline
        $Q$ Hips  &  $a_{2}$:[100, 1500] & $a_{4}$:[10, 300] & $w_{\dot{V}}$ & $a_{6}$:[1, 5] \\
        \hline
    \end{tabular}
\end{table}

% \subsection{Process of Tuning a CLF-QP}
% CLF-QPs have been implemented only on few platforms, in part due to their difficultly in tuning. Mainly, the $Q$ matrix of \eqref{CARE} directly influences the how aggressively the outputs $\eta$ are tracked. A $Q$ matrix with elements that are too small will result in the outputs not be tracked. However, if the elements are too large, torque chatter will be induced into the system.

% The current approach to this tuning problem is to have a domain expert hand tune the elements of $Q$. However, unlike more common controllers such as LQR or traditional PID control, tuning the $Q$ matrix does not have a clear relationship with the final behavior of the system. At first, it appears like the elements of the $Q$ matrix directly correspond with the outputs of the system $\eta$. However, one must keep in mind that $Q$ influences the final controller $u$ indirectly through $P$. Furthermore $P$ influences $u$ indirectly through the CLF $L_F V + L_G V v = - \dot{V}$. Thus, the choice of the $Q$ matrix is further convoluted by the lie derivative over the system dynamics $F$ and $G$. 
%%%%%%%%%%%%%%%%%%%%%%%%%%%%%%%%%%%%%%%%%%%%%%%%%%%%%%%%%%%%%%%%%%%%%%%%%%%%%%%%%%%%%%%%%%%%%%%%%%%%%%%%%%%%%%%%%%%%%%%%%%%%
\subsection{Parameterization of CLF-QP}
% However, the reason to use CLF's despite their difficulty is because they have all of these benefits compared to PD controllers [cite theory papers]. 

For the following discussion, let $\a=[a_1,...,a_v]\in \A \subset \R^v$ be an element of a $v-$dimensional parameter space, termed an \textit{action}. We let $Q=Q(\a)$, $\epsilon = \epsilon(\a)$, and $w_{\dot{V}} = w_{\dot{V}}(\a)$ denote a parameterization of our control tuning variables, which will subsequently be learned. Each gain $a_i$ for $i = 1, \dots, v$ is discretized into $d_i$ values, leading to an overall search space of actions given by the set $\A$ with cardinality $|\A|=\prod_{i=1}^vd_i$. For the AMBER robot, $v$ is taken to be 6 with discretizations $d=[4,4,5,5,4,5]$, resulting in the following parameterization:
\begin{align*}
    Q(\a) &= \begin{bmatrix}Q_1 & 0 \\ 0 & Q_2\end{bmatrix}, ~~~\begin{matrix} Q_1 = \text{diag}([a_1,a_2,a_2,a_1]), \\ Q_2 = \text{diag}([a_3,a_4,a_4,a_3]),\end{matrix} \\
    \epsilon(\a) &= a_5, ~~~~~~~~~~~~~w_{\dot V}(\a) = a_6,
\end{align*}
which satisfies $Q(\a)\succ0$, $0<\epsilon(\a)<1$, and $w_{\dot{V}}(\a) > 0$ for the choice of bounds, as summarized in Table \ref{table:bounds}. Because of the simplicity of AMBER, we were able to tune all associated gains for the CLF-QP$^+$ controller. For Cassie, however, the complexity of the ID-CLF-QP$^+$ controller warranted only a subset of parameters to be selected. Namely, $v$ is taken to be 12 and $d_i$ to be 8, resulting in:
\begin{align*}
    Q &= \begin{bmatrix}Q_1 & 0 \\ 0 & Q_2\end{bmatrix}, ~~~\begin{matrix*}[l] Q_1 =  \text{diag}([a_1,\dots,a_{12}]), \\ Q_2 = \bar{Q},\end{matrix*}
\end{align*}
with $\bar{Q}$, $\epsilon$, and $w_{\dot V}$ remaining fixed and predetermined by a domain expert. From this definition of $Q$, we can split our output coordinates $\eta = (\eta_t,\eta_{nt})$ into \textit{tuned} and \textit{not-tuned} components, where $\eta_t\in \R^{12}$ and $\eta_{nt}\in \R^6$ correspond to the $Q_1$ and $Q_2$ blocks in in $Q$.

\section{Learning Framework} \label{sec:framework}

\begin{figure}[!t]
    \centering
    \includegraphics[width=0.98\linewidth]{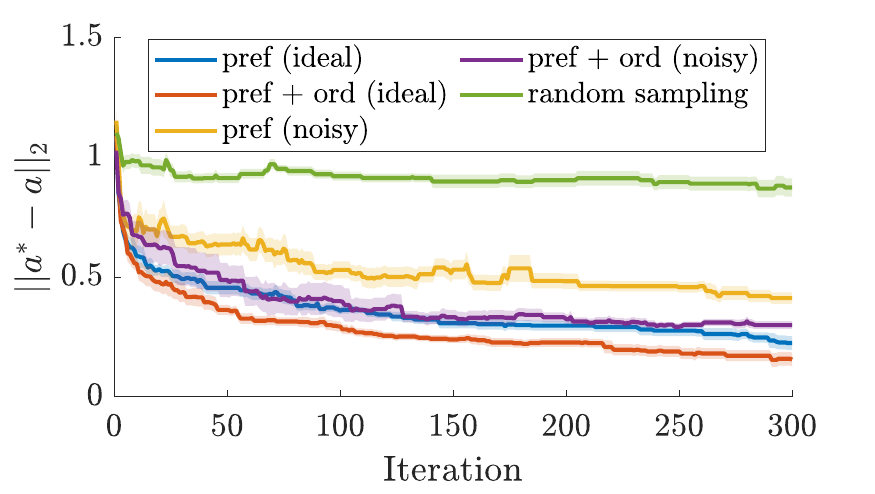}
    \caption{Simulated results averaged over 10 runs, demonstrating the capability of preference-based learning to optimize over large action spaces, specifically the one used for experiments with Cassie. Shaded region depicts standard error.}
    \label{fig:simresults}
\end{figure}

The preference-based learning framework leveraged in this work is a slight extension of that presented in \cite{tucker2020human}. Specifically, this work implements ordinal labels as an additional feedback mechanism to improve sample-efficiency. As in \cite{tucker2020human}, the algorithm is aimed at regret minimization, defined as sampling $N$ actions $\{\a_1,\dots,\a_N\}$ such that:
\begin{align*}
    \{\a_1,\dots,\a_N\} = \argmin_{\a \in \A} \sum_{i = 1}^N \left( \f(\a^*) - \f(\a_i) \right),
\end{align*}
where $\A \subset \R^{|\a|}$ is the discretized set of all possible actions, $\f:\A\to\R$ is the underlying utility function of the human operator mapping each action to a subjective measure of ``good'', and $\a^*$ is the action maximizing $\f$. This objective of regret minimization can be equivalently interpreted as trying to sample actions with high utilities, $\bar{\a}^* = \argmax_{\a \in \A} \f(\a)$, in as few iterations as possible. In this section, we briefly outline the learning framework and how it was modified for our application.

\subsection{Summary of Learning Method}
In each iteration, the user is queried for their preference between the most recently sampled action, $\a_i$, and the previous action, $\a_{i-1}$, denoted as $\a_i \succ \a_{i-1}$ if action $\a_i$ is preferred.
%which we will write as $a_{i-1}$ and $a_{i}$ for iteration $i$. 
% We define a likelihood function based on preferences:
% \begin{align*}
%     \P(a_i \succ a_{i-1} | \f(a_{i}), \f(a_{i-1})) = \begin{cases} 1&\text{if}~ \f(a_{i})\ge \f(a_{i-1})\\0&\text{otherwise},\end{cases}
% \end{align*}
% where $a_i \succ a_{i-1}$ denotes a preference of action $a_i$ over action $a_{i-1}$.
% % for any two compared actions $\{a_1, a_2\} \in \A$. 
% In other words, the likelihood function states that the user has utility $\f(a_i)\ge \f(a_{i-1})$ with probability 1 given that they return a preference $a_{i} \succ a_{i-1}$. 
% %
% % This is a strong assumption about the ability of the user to give noise-free feedback; therefore, to account for noisy preferences we instead use:
% This is a strong assumption on the ability of the user to give noise-free feedback; to account for noisy preferences we instead use:
This preference is modeled as:
\begin{align*}
    \P(\a_i \succ \a_{i-1} | \f(\a_i), \f(\a_{i-1})) = \phi\left( \frac{\f(\a_i)-\f(\a_{i-1})}{c_p} \right),
\end{align*}
where $\phi:\R\to[0,1]$ is a monotonically-increasing link function, and $c_p>0$ represents the amount of noise expected in the preferences. In this work, we select the heavy-tailed sigmoid distribution $\phi(x):=\frac{1}{1+e^{-x}}$.

Inspired by \cite{li2020roial}, we supplement preference feedback with ordinal labels. Because ordinal labels are expected to be noisy, the ordinal categories are limited to only  ``very bad", ``neutral", and ``very good". Ordinal labels are obtained each iteration for the corresponding action $\a_i$ and are assumed to be assigned based on $\f(\a_i)$. 
Similar to preferences, these ordinal labels are modeled using a likelihood function:
\begin{align*}
    \P(o = r | \f(\a_i)) = \phi\left(\frac{b_r-\f(\a_i)}{c_o}\right) - \phi\left( \frac{b_{r-1}-\f(\a_i)}{c_o}\right),
\end{align*}
where $o$ denotes the ordinal label provided by the user with a corresponding ordered ranking $r \in \{1,2,3\}$, $c_0 > 0$ denotes expected noise in the ordinal labels, and $\{b_0, \dots, b_3\}$ are arbitrary thresholds that dictate which latent utility ranges correspond to which ordinal label. 
% In our work, these thresholds were selected to be $\{-\infty, -1, 1, \infty\}$. 

\begin{figure}[t]
	\centering
	\vspace{2mm}
	\includegraphics[width= 0.95\columnwidth]{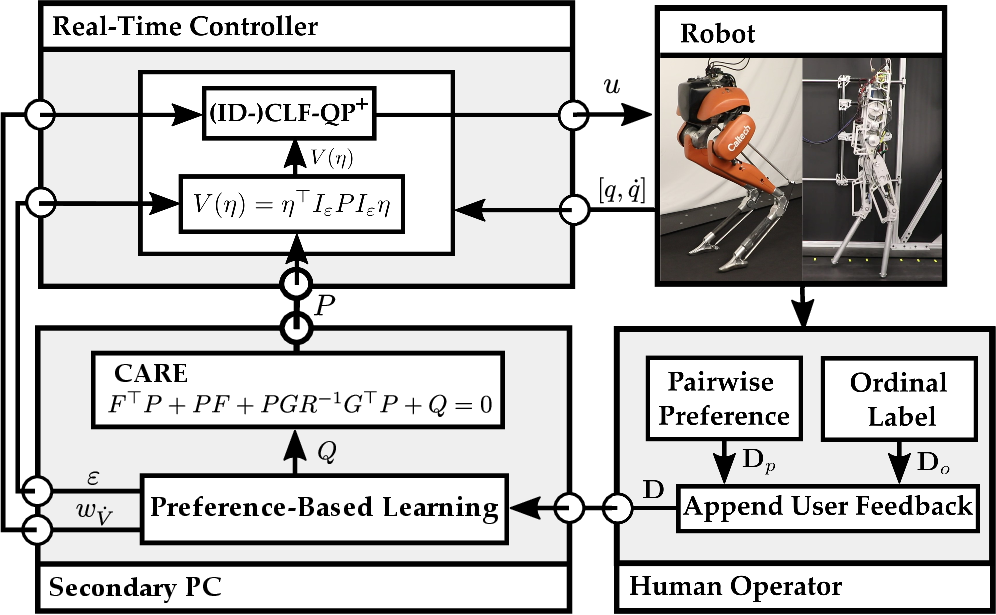}
	\caption{The experimental procedure, notably the communication between the controller, physical robot, human operator, and learning framework.}
	\label{fig:schematic}
\end{figure}

%%%%%%%%%%%%%%%%%%%%%%%%%
\begin{figure*}[t!]
\centering
      \begin{subfigure}[b]{0.45\textwidth}
         \centering
         \includegraphics[width=\linewidth]{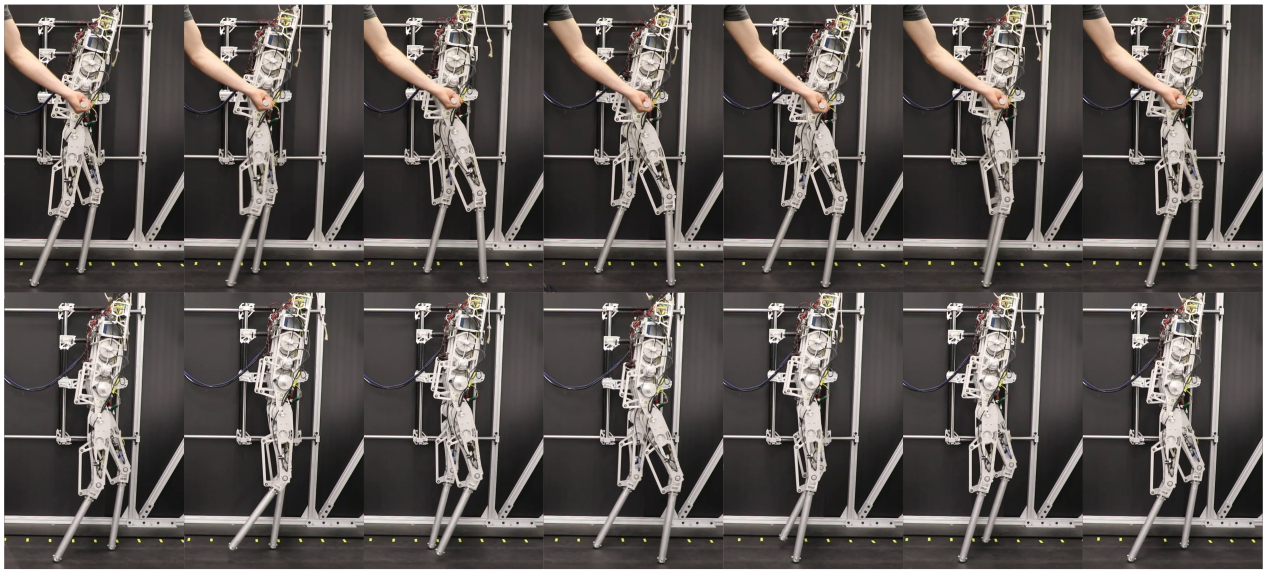}
         \caption{Very low utility (top) where the robot was unable to walk unassisted and maximum posterior utility (bottom) where stable walking was achieved.}
         \label{fig:AMBERgaitTiles}
     \end{subfigure}
     \hspace{3mm}
     \begin{subfigure}[b]{0.45\textwidth}
         \centering
         \includegraphics[width=\linewidth]{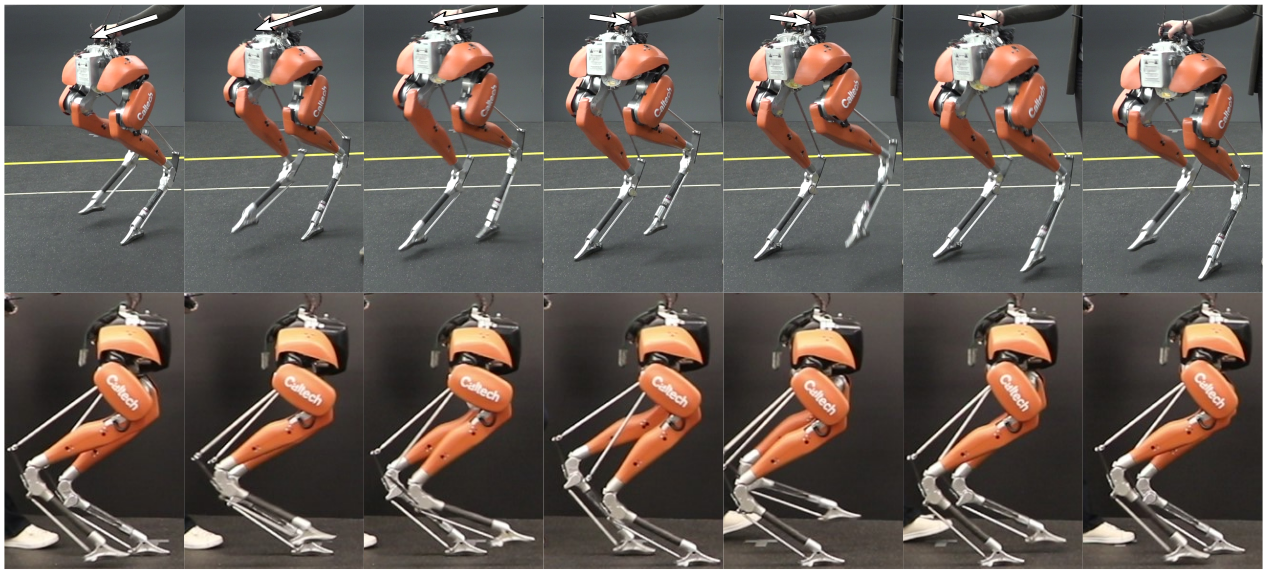}
         \caption{The robustness (top) and tracking (bottom) of the walking with the learned optimal gains, enabling unassisted, stable walking and good tracking performance.}
         \label{fig:tiles}
     \end{subfigure}
     \caption{Gait tiles for AMBER (left) and Cassie (right).}
    \end{figure*}
    \begin{figure*}[t!]
    \centering
     \begin{subfigure}[b]{0.49\textwidth}
         \centering
         \includegraphics[width=\linewidth]{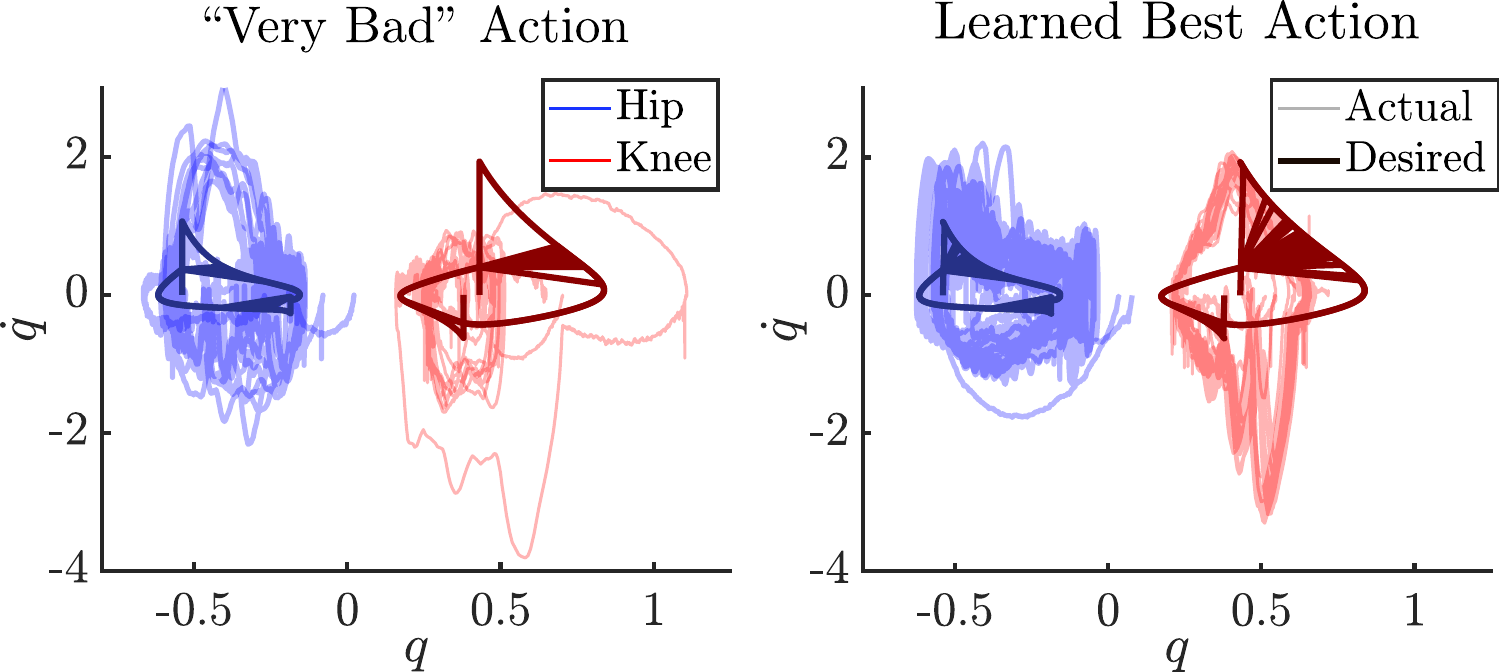}
         \caption{Phase portraits for AMBER experiments.}
         \label{fig:phaseAMBER}
     \end{subfigure}
          \begin{subfigure}[b]{0.49\textwidth}
         \centering
         \includegraphics[width=\linewidth]{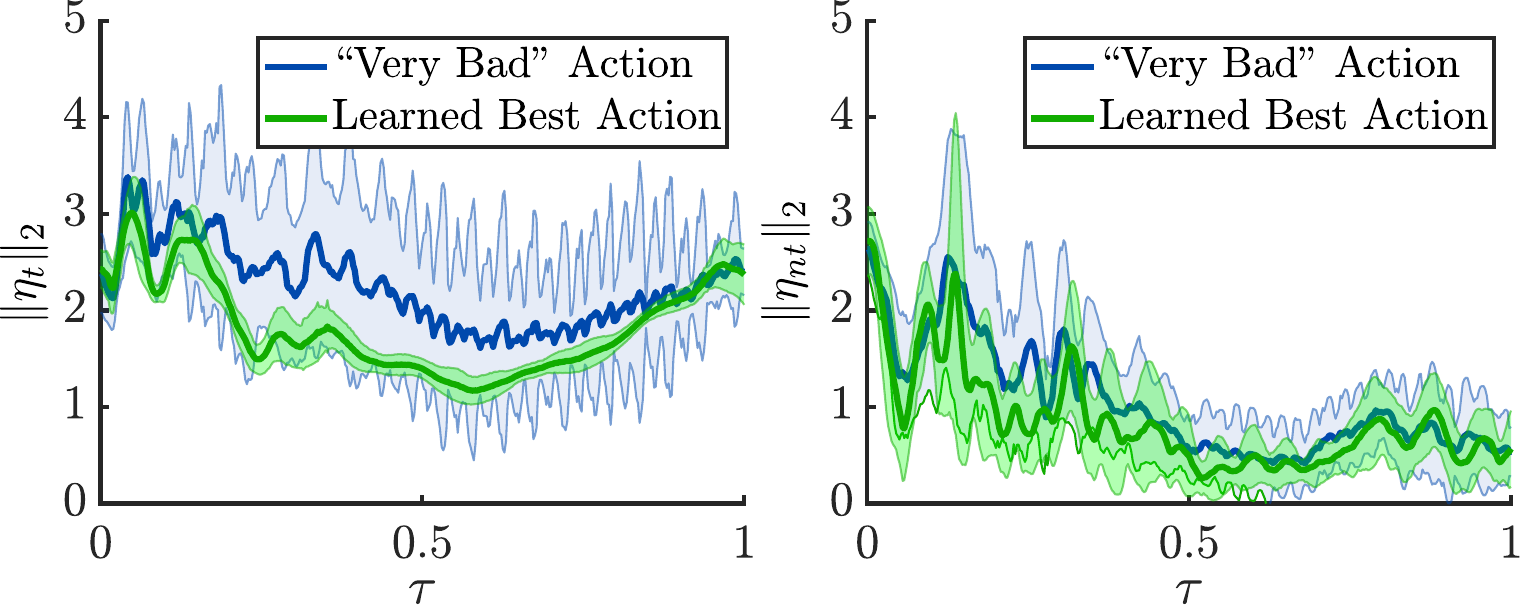}
         \caption{Output Error of $\eta_{t}$ (left) and $\eta_{nt}$ (right) for Cassie experiment.}
         \label{fig:eta}
     \end{subfigure}
         \caption{Experimental walking behavior of the CLF-QP$^+$ (left) and the ID-CLF-QP$^+$ (right) with the learned gains.}
         \label{fig:expplots}
         \vspace{-4mm}
\end{figure*}

In each iteration, operator feedback is obtained and appended to the preference and ordinal label datasets $\D_p$ and $\D_o$, with all feedback denoted as $\D = \D_p \cup \D_o$. 
% This feedback is then used to infer the latent utilities of the sampled actions  by applying the preference-based Gaussian process model to the posterior  distribution $\P(\fvec \mid \D)$ as in \cite{PrefLearn}.
This feedback is then used to approximate the posterior distribution $\P(\fvec\mid\D)$ as a multivariate Gaussian $\mathcal{N}(\mu,\Sigma)$ via the Laplace approximation as in [\citenum{PrefLearn}, Sec. 2.3]. To remain tractable in high-dimensions, $\fvec$ is a restriction of $\f$ as defined in \cite{tucker2020human}. The predictive distribution at location $\hat{\a}$ is described by a univariate Gaussian $\f(\hat{\a}) \sim \mathcal{N}(\mu(\hat{\a}),\Sigma(\hat{\a}))$, whose equations can be found in [\citenum{PrefLearn}, Sec. 2.3].

% latent utilities of the sampled actions $\fvec = [\f(\a_1),\ldots,\f(\a_{N})]^\top$ by applying the preference-based Gaussian process model to the posterior distribution $\P(\fvec \mid \D)$ as in \cite{PrefLearn}. 
% In general, this approach yields an approximation of $\P(\fvec \mid \D)$ as a multivariate Gaussian $\mathcal{N}(\mu,\Sigma)$. 
% The mean $\mu$ can be interpreted as our estimate of the latent utilities $\hat{\fvec}$, with $\hat{\a}^* = \argmax_{\a \in \A} \mu(\a)$ being the estimated optimal action after the most recent update.

% First, we model the posterior distribution as proportional to the likelihoods multiplied by the Gaussian prior using Bayes rule,
% \begin{align}
%     \P(\fvec|\D_p,\D_o) \propto \P(\D_o,\D_p|\fvec)\P(\fvec), \label{post}
% \end{align}
% where the Gaussian prior over $\fvec$ is given by:
% \begin{align*}
%     \P(\fvec) = \frac{1}{(2\pi)^{|\V|/2}|\Sigma|^{1/2}}\exp\left(-\frac{1}{2}\fvec^\top \Sigma^{-1}\fvec\right).
% \end{align*}
% with $\Sigma\in\R^{|\V|\times |\V|}$, $\Sigma_{ij} = \mathcal{K}(a_i,a_j)$, and $\mathcal{K}$ is a kernel.
% Assuming conditional independence of queries, we can split $\P(\D_o,\D_p|\fvec)=\P(\D_o|\fvec)\P(\D_p|\fvec)$ wherse
% \begin{align*}
%     \P(\D_p|\fvec) &= \prod_{i=1}^K\P(a_1\succ a_2 | \f(a_1),\f(a_2)),\\
%     \P(\D_o|\fvec) &= \prod_{i=1}^M\P(a_1=r_1 | \f(a_1)).
% \end{align*}
% The posterior \eqref{post} is then estimated via the Laplace approximation as in \cite{PrefLearn},

To select new actions to query in each iteration, Thompson sampling \cite{chapelle2011empirical} is used. Specifically, at each iteration, a function $\hat\f$ randomly drawn from the Gaussian process is maximized. 
% The idea of Thompson sampling is to select, at each iteration, the action that maximizes the sample  $\hat\f$ as:
% Finally, new actions are selected for the subsequent iteration via Thompson sampling. For each sampled action, this sampling method draws a random sample from $\f \sim \mathcal{N}(\mu,\Sigma)$ and sets the action which maximizes $\f$. 
This iterative process of (1) querying the operator for feedback, (2) modeling the underlying utility function, and (3) sampling new actions, is repeated in each subsequent iteration. Finally, the best action after the completion of the experiment is given by $\hat\a^* = \argmax_{\a \in \A} \mu(\a)$.
\begin{table}[b!]
\centering
\caption{Learned Parameters}
\label{table:optimal}
\def\arraystretch{1.2} % 1 is the default, increase for more row spacing
        \begin{tabular}{ | c | c |} 
        \hline
        % \multicolumn{1}{|c|}{\textsc{CASSIE}} \\
        % \hline
        AMBER & $[750, 100, 300, 100, 0.125, 2]$\\
        \hline
        % \multicolumn{1}{|c|}{\textsc{AMBER}} \\
        % \hline
        Cassie &$[2400,1700,4200,5600, 1700,1200,27,40,120,56,17,7]$\\
        \hline
    \end{tabular}
\end{table}
\subsection{Expected Learning Behavior}
To demonstrate the learning, a simple example was constructed of the same dimensionality as the parameter space being investigated on Cassie ($v = 12, d = 8$), where the utility was modeled as $\f(\a) = \|\a-\a^*\|_2$ for some $\a^*$. Feedback was automatically generated for both ideal noise-free feedback as well as for noisy feedback (correct feedback given with probability 0.9). The results of the simulated algorithm, illustrated in Fig. \ref{fig:simresults}, show that the learning framework quickly samples actions near $a^*$, even for an action space as large as the one used in the experiments with Cassie. The simulated results also show that ordinal labels 
% allow for faster 
improve convergence,
% to the optimal action, even in the case of noise,
motivating their use in the final experiment. 
% In comparison, the random sampling method leads to minimal improvement when compared to preference-based learning. From these simulation results, it can clearly be seen that the proposed method is an effective mechanism for exploring high-dimensional parameter spaces. 

\section{Learning to Walk in Experiments} \label{sec:exp}
Preference-based learning was applied to the realization of optimization-based control on two separate robotic platforms: the 5 DOF planar biped AMBER, and the 22 DOF 3D biped Cassie, as can be seen in the video \cite{video}. 
%
% Next, the preference-based learning framework, applied to the realization of optimization-based control, is demonstrated on two separate robotic platforms: the 5 DOF planar biped AMBER, and the 22 DOF 3D biped Cassie, as can be seen in the video \cite{video}. 
%
As illustrated in Fig. \ref{fig:schematic}, the experimental procedure had four main components: the physical robot (either AMBER or Cassie), the controller running on a real-time PC, a human operator providing feedback, and a secondary PC running the learning algorithm. 
% was tasked with providing the user feedback based on their perspective of the robot behavior
Each action was tested for approximately one minute, during which the behavior of the robot was evaluated in terms of both performance and robustness. User feedback in the form of pairwise preferences and ordinal labels was obtained after testing each action via the respective questions: ``Do you prefer this behavior more or less than the last behavior'', and ``Would you give this gait a label of very bad, neutral, or very good''. After user feedback was collected for the sampled controller gains, the posterior was inferred over all of the uniquely sampled actions, which took up to 0.5 seconds. The experiment with AMBER was conducted for 50 iterations, lasting one hour, and the experiment with Cassie was conducted for 100 iterations, lasting two hours. The duration of the experiments was scaled based on the size of the respective action spaces, and trials were terminated when satisfactory behaviors had been sampled.

% The number of iterations conducted differed depending on the robot used: the experiment with AMBER was conducted for 50 iterations, the experiment with Cassie was conducted for 100 iterations. 
% This is because the action space for Cassie was much larger than that for AMBER.
% The experiment with AMBER lasted approximately one hour, and the experiments with Cassie lasted roughly one hour for the domain expert and roughly two hours for the na\"ive user. 
% The reason why the 100 iteration experiments with Cassie lasted less than two hours was because the gains were updated continuously so as to facilitate faster feedback. 

% \red{Something about the reason to do AMBER first and then follow with Cassie - need to hit home the transition from simple to complicated}

\begin{figure*}[t!]
\centering
         \centering
         \includegraphics[width=0.96\linewidth]{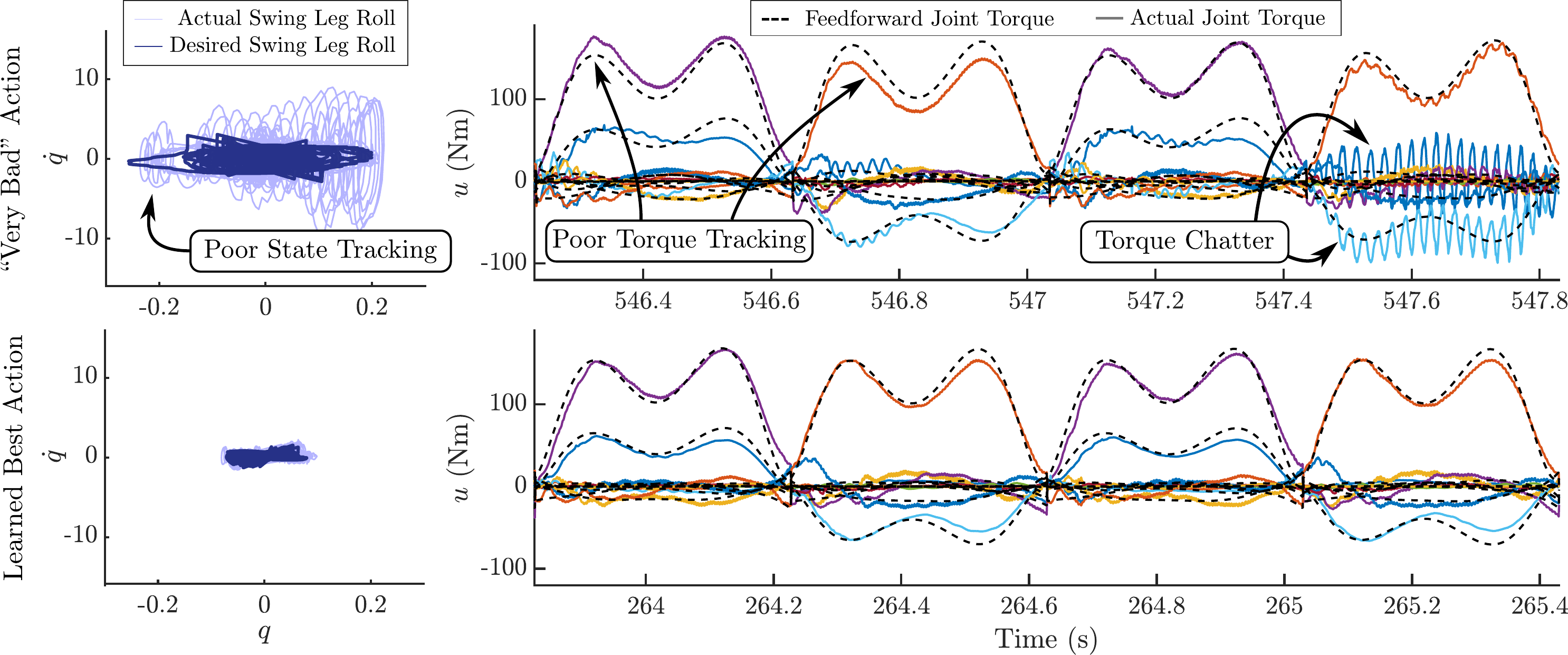}
         \caption{Phase plots and torques commanded by the ID-CLF-QP$^{+}$ in the na\"ive user experiments with Cassie. For torques, each colored line corresponds to a different joint, with the black dotted lines being the feedforward torque. The gains corresponding to a ``very bad'' action (top) yield torques that exhibit poor tracking on joints and torque chatter. On the other hand, the gains corresponding to the learned optimal action (bottom) exhibit much better tracking and no torque chatter.}
         \label{fig:torque}
    %  \begin{subfigure}[b]{0.47\textwidth}
    %      \centering
    %      \includegraphics[width=\linewidth]{Figures/worstTorque.pdf}
    %      \caption{Controller torques corresponding to a ``Very Bad'' action.}
    %      \label{fig:torque_bad}
    %  \end{subfigure}
    %  \hfill
    %  \begin{subfigure}[b]{0.47\textwidth}
    %      \centering
    %      \includegraphics[width=\linewidth]{Figures/bestTorque.pdf}
    %      \caption{Controller torques corresponding to the learned best gains.}
    %      \label{fig:torque_good}
    %  \end{subfigure}
    %      \caption{The torques commanded by the ID-CLF-QP$^{+}$ in the na\"ive user experiments with Cassie are plotted for a single step. Each colored line corresponds to a different joint, with the black dotted lines being the feedforward torque. The gains corresponding to a ``very bad'' action yield torques (left) that exhibit poor tracking and torque chatter. On the other hand, the gains corresponding to the learned optimal action (right) exhibit much better tracking and no torque chatter.}
    %      \label{fig:torques}
         \vspace{-4mm}
\end{figure*}

\subsection{Results with AMBER -- CLF-QP$^+$}
\label{sec: AMBERexp}
The CLF-QP$^+$ controller was implemented on an off-board i7-6700HQ CPU @ 2.6GHz with 16 GB RAM, which solved for desired torques and communicated them with the ELMO motor drivers on the AMBER robot at 2kHz.
% The action space of control parameter combinations was defined using the bounds reported in Table \ref{table:bounds}. Once the action space was defined, 50 iterations of the algorithm were run in an hour.
During the first half of the experiment, the algorithm sampled a variety of gains causing behavior ranging from instantaneous torque chatter to induced tripping due to inferior output tracking. It is important to note that none of the initial sampled values led to unassisted walking.
% Empirically, when $Q$ was large in magnitude, $\epsilon$ was driven too low, and $w_{\dot V}$ was too high, the QP went unstable and \
By the end of the experiment however, the algorithm had sampled 3 gains which were deemed "very good", and which resulted in stable walking. The final learned best actions found by the algorithm are reported in Table \ref{table:optimal}. Gait tiles for an action deemed ``very bad", as well as the learned best action are shown in Fig. \ref{fig:AMBERgaitTiles}. Additionally, tracking performance for the two sets of gains is seen in Fig. \ref{fig:phaseAMBER}, where the learned best action tracks the desired behavior to a better degree.

\subsection{Results with Cassie -- ID-CLF-QP$^+$}
\label{sec: Cassieexp}
% To test the capability of the learning method towards tuning more complex controllers, the preference-based learning method was applied for tuning the gains of the ID-CLF-QP$^+$ controller for the Cassie bipedal robot. 
The ID-CLF-QP$^+$ controller was implemented on the on-board Intel NUC computer, which was running a PREEMPT\_RT kernel. The software runs on two ROS nodes, one of which communicate state information and joint torques over UDP to the Simulink Real-Time xPC, and one of which runs the controller. Each node is given a separate core on the CPU, and is elevated to real-time priority. Preference-based learning was run on an external computer and was connected to the ROS master over WiFi. Actions were updated in real-time; once an action was selected, it was sent to Cassie via a rosservice call, where, upon receipt, the robot immediately updated the corresponding gains. As rosservice calls are blocking, multithreading their receipt and parsing was necessary in order to maintain real-time performance.

To demonstrate repeatability, the experiment was conducted twice on Cassie: once with a domain expert, and once with a na\"ive user. In both experiments, a subset of the $Q$ matrix from $\eqref{CARE}$ was tuned with coarse bounds given by a domain export, as reported in Table \ref{table:bounds}. These specific outputs were chosen because they were deemed to have a large impact on the performance of the controller. 
Some metrics used to determine preferences were the following: no torque chatter, no drift in the floating base frame, responsiveness to desired directional input, no violent impacts, no harsh noise, and naturalness of walking. At the start of the experiments, there was significant torque chatter and wandering, with the user having to regularly intervene to recenter the global frame. As the experiments continued, the walking noticeably improved. At the conclusion of 100 iterations, the posterior was inferred over all uniquely visited actions. The action corresponding with the maximum utility -- believed by the algorithm to result in the most user preferred walking behavior -- was further evaluated for tracking and robustness. In the end, this learned best action coincided with the walking behavior that the user preferred the most. 
% The optimal gains identified by the framework are: 

% \vspace{-3mm}
% {\small \begin{align*}
% \hat{a}^* = &[2400,1700,4200,5600, 1700,1200,27,40,120,56,17,7]. \notag
% \end{align*}}
% \vspace{-3mm}

Features of this optimal action, compared to a worse action sampled in the beginning of the experiments, are outlined in Fig. \ref{fig:expplots}. In terms of quantifiable improvement, the difference in tracking performance is shown in Fig. \ref{fig:eta}. 
% For the sake of presentation, the outputs are split into $\eta = (\eta_t,~\eta_{nt})$ where $\eta_t$ are the 12 outputs whose parameters were tuned by the learning algorithm and $\eta_{nt}$ are the remaining 6 outputs.
The magnitude of the tuned parameters, $\eta_t$, illustrates the improvement that preference-based learning attained in tracking the outputs it intended to. At the same time, the tracking error of the constant parameters, $\eta_{nt}$, shows that the outputs that were not tuned remained unaffected by the learning process. This quantifiable improvement is further illustrated by the commanded torques in Fig. \ref{fig:torque}, which show that the optimal gains result in much less torque chatter and better tracking as compared to the other gains.
% \red{talk about why terms were coupled in AMBER for Q}

\subsection{Limitations and Future Work}
The main limitation of the current formulation of preference-based learning is that the action space must be predefined with set bounds. In the context of controller gains, these bounds are difficult to know \textit{a priori} since the relationship between the gains and the resulting behavior is unpredictable. Future work to address this problem involves modifications to the learning framework to shift action space based on the user's preferences. Furthermore, the current framework limits the set of potential new actions to the set of actions discretized by $d_i$ for each dimension $i$. As such, future work also includes adapting the granularity of the action space based on the uncertainty in specific regions. 

% In other words, as the algorithm becomes more confident that the optimal action is located in a coarse region, the algorithm could explore a finer grained action space within this region. 
%%%%%%%%%%%%%%%%%%%%%% Moderate line %%%%%%%%%%%%%%%%%%%%%%%%%%%%%%%%%%%%%%%%

\section{Conclusion} 
\label{sec:conc}
Navigating the complex landscape of controller gains is a challenging process that often requires significant knowledge and expertise. In this work, we demonstrated that preference-based learning is an effective mechanism towards systematically exploring a high-dimensional controller parameter space, without needing to define an objective function. Furthermore, we experimentally demonstrated the power of this method on two different platforms with two different controllers, showing the application agnostic nature of the framework. In all experiments, the robots went from stumbling to walking in a matter of hours. Additionally, the learned best gains in both experiments corresponded with the walking trials most preferred by the human operator. In the end, the robots had improved tracking performance, and were robust to external disturbance. Future work includes addressing the aforementioned limitations, extending this methodology to other robotic platforms, coupling preference-based learning with metric-based optimization techniques, and addressing multi-layered parameter tuning tasks.

% *********************************************** 

% \section*{ACKNOWLEDGMENTS}
% The authors would like to thank the sponsors who made this work possible.

% \newpage 
\bibliographystyle{IEEEtran}
\balance
\bibliography{IEEEabrv,References}

\end{document}